\documentclass[conference,compsoc]{IEEEtran}
\IEEEoverridecommandlockouts
\usepackage{cite}
\usepackage{amsmath,amssymb,amsfonts}
\usepackage{algorithmic}
\usepackage{graphicx}
\usepackage{textcomp}
\usepackage{xcolor}

\def\BibTeX{{\rm B\kern-.05em{\sc i\kern-.025em b}\kern-.08em
    T\kern-.1667em\lower.7ex\hbox{E}\kern-.125emX}}

\newcommand{\IR}{\mathbb{R}}           

\renewcommand{\epsilon}{\varepsilon}                    
\newcommand{\abs}[1]{\left\lvert #1 \right\rvert}       
\newcommand{\norm}[1]{\left\lVert #1 \right\rVert}      

\newcommand{\sample}{x}

\newcommand{\labelentry}{y}
\newcommand{\bias}{b}

\newcommand{\optomega}{\tilde{\omega}}
\newcommand{\optomeganorm}{\onenorm{\optomega{}}}

\newcommand{\onenorm}[1]{\norm{#1}_1}

\newcommand{\regparameter}{\mathit{C}}
\newcommand{\slack}{\xi}
\newcommand{\optslack}{\tilde{\slack}}
\newcommand{\optoffset}{\tilde{b}}
\newcommand{\costsum}[1]{\ensuremath{\sum_{i=1}^{n} \slack{}_i}}
\newcommand{\losssum}[1]{\regparameter \cdot \sum_{i=1}^{n} #1_i}

\newcommand{\data}{\mathbb{D}}

\DeclareMathOperator*{\argmin}{arg\,min}

\newcommand{\sone}{\emph{Sim1}}
\newcommand{\stwo}{\emph{Sim2}}
\newcommand{\sthree}{\emph{Sim3}}
\newcommand{\sfour}{\emph{Sim4}}
\newcommand{\sfive}{\emph{Sim5}}

\newcommand{\fri}{\emph{FRI}}
\usepackage{flushend}
\IEEEoverridecommandlockouts
\IEEEpubid{\makebox[\columnwidth]{978-1-7281-1462-0/19/\$31.00~\copyright{}2019 IEEE \hfill}\hspace{\columnsep}\makebox[\columnwidth]{ }}
\begin{document}

\title{FRI - Feature Relevance Intervals for Interpretable and Interactive Data Exploration\\
\thanks{This contribution has been made possible through
the German DFG research training group 
“Computational Methods for the Analysis of the Diversity and Dynamics of Genomes” 
(DiDy) GRK 1906/1.}
}

\author{
    \IEEEauthorblockN{1\textsuperscript{st} Lukas Pfannschmidt}
    \IEEEauthorblockA{\textit{DiDy \& Machine learning group} \\
    \textit{Bielefeld University}\\
    Bielefeld, Germany\\
    lpfannschmidt@techfak.uni-bielefeld.de}
    \and
    \IEEEauthorblockN{2\textsuperscript{nd} Christina Göpfert}
    \IEEEauthorblockA{\textit{Machine learning group} \\
    \textit{Bielefeld University}\\
    Bielefeld, Germany\\
    cgoepfert@techfak.uni-bielefeld.de}
    \and
    \IEEEauthorblockN{3\textsuperscript{rd} Ursula Neumann}
    \IEEEauthorblockA{\textit{Dep. of Mathematics and Computer Science} \\
    \textit{University of Marburg}\\
    Marburg, Germany\\
    ursula.neumann@staff.uni-marburg.de}
    \and
    \IEEEauthorblockN{4\textsuperscript{th} Dominik Heider}
    \IEEEauthorblockA{\textit{Dep. of Mathematics and Computer Science} \\
    \textit{University of Marburg}\\
    Marburg, Germany\\
    dominik.heider@uni-marburg.de}
    \and
    \IEEEauthorblockN{5\textsuperscript{th} Barbara Hammer}
    \IEEEauthorblockA{\textit{Machine learning group} \\
    \textit{Bielefeld University}\\
    Bielefeld, Germany\\
    bhammer@techfak.uni-bielefeld.de}
}

\maketitle

\begin{abstract}
Most existing feature selection methods are insufficient for analytic purposes as soon as high dimensional data or redundant sensor signals are dealt with since features can be selected due to spurious effects or correlations rather than causal effects.
To support the finding of causal features in biomedical experiments, we hereby present FRI, an open source \emph{Python} library that can be used to identify all-relevant variables in linear classification and (ordinal) regression problems.
Using the recently proposed \emph{feature relevance interval} method, FRI is able to provide the base for further general experimentation or in specific can facilitate the search for alternative biomarkers.
It can be used in an interactive context, by providing model manipulation and visualization methods, or in a batch process as a filter method.
\end{abstract}


\section{Background}
  \label{sec:intro}

  In recent years, due to an increasing availability of highly sensitive biotechnologies as well as an increasing digitalization of biomedical diagnostics, one could observe a trend towards bigger and more complex machine learning models.
  These models are used successfully in medical diagnoses \cite{kononenko_machine_2001}\cite{bellazzi_predictive_2008} such as cancer prediction \cite{cruz_applications_2006-1} using known biomarkers as input.
  When specific biomarkers are unknown, many learning models can also perform on nearly raw data without a preselection of variables.
  While such data representation often enables a high prediction accuracy, it is less suited if the purpose is data exploration and understanding of the underlying causal relationships. 
  For the latter, insight into the model behavior and its relevant driving factors is necessary \cite{vellido_alcacena_making_2012} and feature selection constitutes a first step to  unravel the underlying relationships.
  But even for predictive models, the inference of sparse models can have a significant impact on the model performance, since it helps to mediate the curse of dimensionality thus leading to a better generalization ability and improved computational complexity.
  Because of this, there is a big need for methods to reveal relevant structure and function in the data itself.

  Feature Selection (FS) constitutes one particularly prominent paradigm which enables the inference of sparse and interpretable prediction models \cite{guyon_introduction_2003}.
  The task is mostly undertaken in conjunction with model selection to improve predictive performance.
  Then the problem is often defined as finding the minimal subset of all features to achieve the best performance given some objective score.
  Commonly, one distinguishes filter, wrapper, and embedded methods.
  Filter methods are based on the information content of features as regards the output class label, disregarding the specific classification method.
  Hence they are particularly suited as a first screening technology for high dimensional data \cite{Yu:2003:FSH:3041838.3041946}.
  Wrapper approaches add (remove) features to a growing (shrinking) candidate set~\cite{kohavi_wrappers_1997} while evaluating an inner model on a predefined metric.
  Embedded approaches use the internal model weights to find relevant features such as the \emph{Lasso}~\cite{tibshirani_regression_1996} which enforces sparsity through its choice of regularization.
  Since they do not rely on iterative feature selection or weighting, embedded approaches have the benefit that they can effectively take into account interdependencies of groups of features. Further, they are specific to the used model. 

  In this contribution, we will focus on embedded methods for one particularly prominent group of models, namely linear classifiers or regression models, such as also addressed by the famous \emph{Lasso}.
  While linear models can not applied to every problem, they are widely used in medicine \cite{ApplicationsSupportVector2018} and the growing size of data makes their efficiency even more relevant.
  Furthermore, methods such as \emph{Lasso} can be accompanied by formal guarantees on their ability to identify the true underlying features in the limit \cite{zhao_model_2006}.
  Yet, these guarantees do not hold if conditions as regards unique representability  are violated.
  The latter is the case if features are high dimensional, highly correlated, or if there do exist different feature sets which yield the same prediction accuracy.
  In such cases, prediction accuracy in the limit still holds \cite{greenshtein_persistence_2004} yet the fact that \emph{Lasso} can identify the true underlying features is violated.
  In particular, greedy and sparse approaches lead to instabilities of the selected feature set when correlated features are present.
  They remove similar features which could otherwise be grouped into potential functional units.

  Finding a minimal subset (the usual objective of \emph{Lasso} or related methods) is therefore not suited for giving a complete picture of all relevant features but only those which are sufficient to fulfill the prediction.
  To gain information about the importance of \emph{all} features one needs to consider a more general measure: \emph{feature relevance}\cite{kohavi_wrappers_1997}\cite{BellFormalismRelevanceIts}.
  The membership of the minimal feature set as a binary measure of relevance is too narrow in most cases.
  \emph{Kohavi et al.}~\cite{kohavi_wrappers_1997} coined the term \emph{all relevant feature selection} (ARFS) in the 90s and added further distinction of relevance.
  Additional to strongly relevant features, which need to be part of the candidate feature set to achieve the best performance, and irrelevant features which are not beneficial for the considered relationship at all, they also introduced the concept of weakly relevant features i.e. features which can contribute information to a model, but are not necessarily included in every good model.

  Weakly relevant features are often highly correlated with each other.
  They are not limited to pairs but also bigger groups of associated features.  
  Out of these, at least one has to be part of the candidate set.
  Having knowledge about a group of features which could fulfill the same role in a model can be very important in the design of diagnostic tests where the source of data can differ by the cost or invasiveness of acquisition.
  Explicit redundancies of feature relevance then enable a practitioner to avoid, e.g., expensive features if they can be substituted by others.
  Additionally, feature groups could induce novel biological relationships.
  This is especially useful in gene co-expression or metabolomics experiments where groups of functional units are common~\cite{van_dam_gene_nodate}.
  Interestingly, it has been shown that extensions of \emph{Lasso} which take into account feature correlations such as group \emph{Lasso} can lead to wrong results due to a selection bias caused by feature correlations~\cite{tolosi_classification_2011}.

  Finding an optimal solution to the ARFS problem is computationally intractable~\cite{kumar_feature_2014} but approximations exist.
  In 2005 the \emph{ElasticNet}~\cite{zou_regularization_2005} overcame some of the instability of sparse models~\cite{Lecun1995Learning} by using a combination of multiple different regularization terms which lead to better conservation of weakly relevant features.
  The \emph{statistically equivalent signature (SES)}~\cite{lagani_feature_2016} approach proposes a technology which groups mutually equivalent features into groups out of which minimal feature subsets can be constructed.
  Another proposal called \emph{stability selection} uses resampling for more robust selection especially in high dimensional problems\cite{MeinshausenStabilityselection2010}\cite{ShahVariableselectionerror2013}.
  In 2010 \emph{Boruta}~\cite{Kursa2010Feature} specifically focused on finding all relevant features using statistical testing of contrast features.
  Alternatives to the Boruta method are discussed and evaluated in \cite{degenhardt_evaluation_2017}, whereby Boruta was identified as best performing technology among the tested ones if used for different dimensionalities of the data.
  In 2017 \emph{Neumann et al.} presented the \emph{Ensemble Feature Selection} (EFS)
  method which combines multiple FS methods to remove individual biases and give aggregated feature relevance ranges for all of them  \cite{neumann_efs_2017, neumann_compensation_2016}.

  In this paper, we focus on the question of how to efficiently uncover a detailed view on the relevance of features in the case of possible feature redundancies.
  Thereby, we investigate linear models as a particularly relevant setting.
  The main contribution of this article is an accessible implementation of the \emph{feature relevance interval} method (\fri{})~\cite{gopfert_interpretation_2018}\footnote{Source code and package available at github.com/lpfann/fri.}.
  The \fri{} offers an efficient framework which assigns relevance intervals to features, rather than simple coefficient values. 
  These intervals mirror the coefficient range of a feature when considering \emph{all} possible models, hence offering detailed and complete information also in the case of feature redundancies.
  In this contribution, we offer an interactive software tool based on the mathematical framework as derived in \cite{gopfert_interpretation_2018} and we demonstrate its applicability in the context of biomedical data analysis.
  More precisely, we show that using these bounds, we can classify each feature into one of the three relevance classes.
  Furthermore, we can use these relevance bounds in visualizing the model which allows interpretability. 
  Especially the information provided by the discrimination between strongly and weakly relevant features can help in biomedical applications.
  As an example we look at model design and biomarker discovery, where it could highlight elements which are crucial for the problem at hand while also providing alternative markers which have the same information.

\begin{figure*}[tb]
        \centering
        \includegraphics[width=0.7\textwidth]{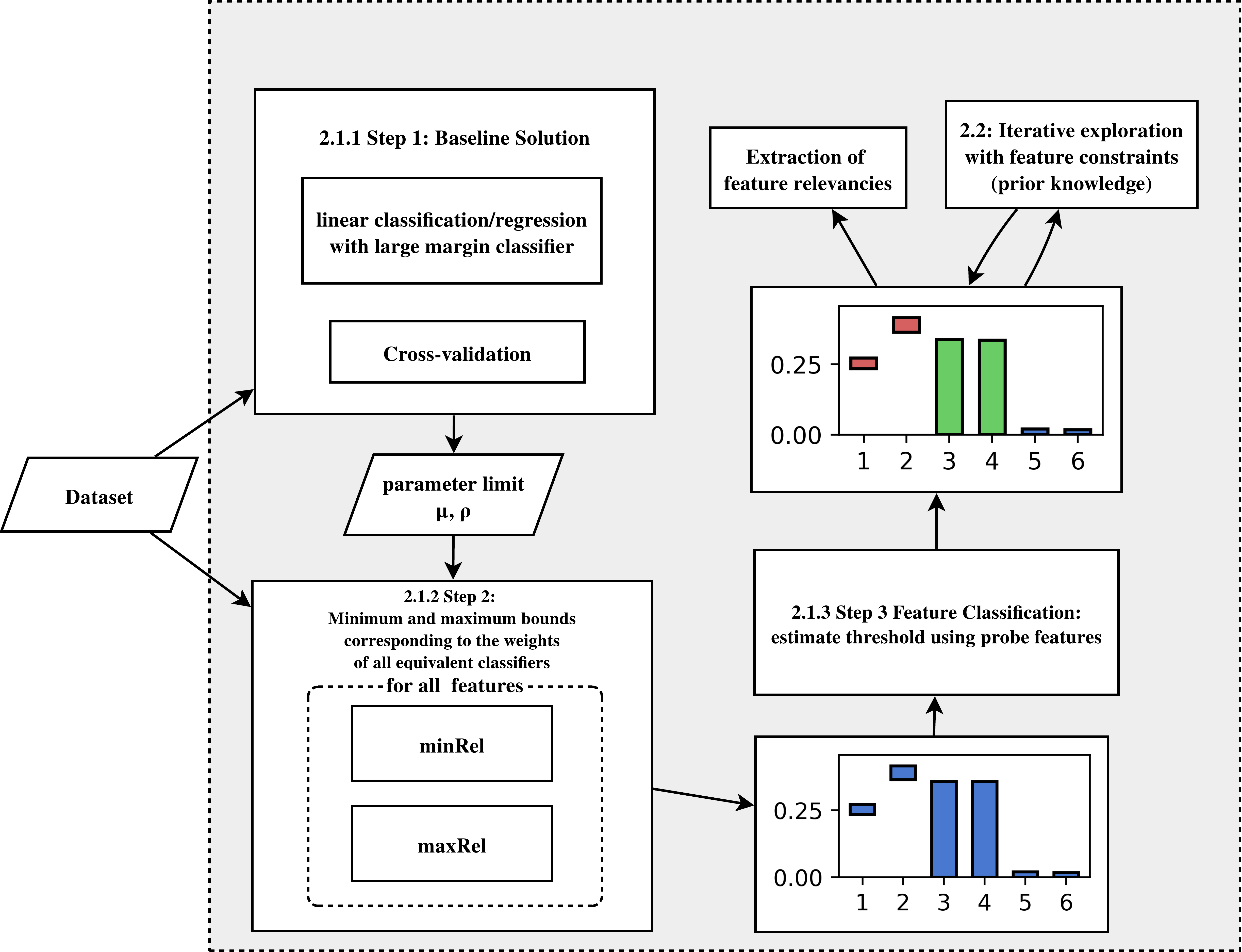}
        \caption{Overview of \fri.}
        \label{fig:pipeline}
      \end{figure*}

  Figure~\ref{fig:pipeline} displays the structure of our proposed software pipeline and in part this article. To elaborate: in Chapter~\ref{sec:implem} we shortly recapitulate the theoretical background and give details of the implementation of the method using linear programming.
  Specifically in \ref{ssub:baseline_solution} we show how to obtain a baseline solution and give our definition of relevance bounds in \ref{ssub:relevance_bounds}.
  We then describe how we can classify each feature into three relevance groups and how to reduce false positives using a probe based threshold estimation in \ref{ssub:feature_classification}.
  Then we show how we can constrain the use of features to certain relevance values (\ref{sub:feature_constraints}) to facilitate interactive data exploration and model design.
  In \ref{ssub:evaluation} we evaluate the method quantitatively using simulated (\ref{sub:simulation_data}) and biomedical data (\ref{sub:biomedical_data}).
  Finally we also provide an application focused qualitative evaluation and example in~\ref{sub:example_of_interactive_use}.
\section{Implementation}
  \label{sec:implem}

    \subsection{Feature Relevance Intervals} 
      \label{sub:relevance_optimization}
      For brevity, in the following definitions and experiments we only consider the task of binary classification, but the method and our implementation can be applied to linear regression and ordinal regression problems as well \cite{PfannschmidtFeatureRelevanceBounds2019}.

      For a classification problem we observe data $\data$ as $
        (x_1,y_1),\dots,(x_n,y_n)\ \in \mathbf{R}^d \times \{-1,1\}
      $
      with $n$ different samples and $d$ real-valued features which have been tied to a target or response $y$.
      We assume that all $d$ features have been standardized at mean zero and standard deviation~1.

      We propose an interactive pipeline, which enables the practitioner a detailed view of the relevance of features on the classification also in the case of multiple solutions or feature redundancies.
      More precisely, we propose an efficient framework which determines the relevance of features as regards all possible solutions of the classification problem, we demonstrate how this information can be used to identify all strongly as well as weakly relevant features, and we present a procedure to iteratively investigate feature combinations.

      \subsubsection{Baseline Solution} 
      \label{ssub:baseline_solution}
      It is common practice to evaluate the relevance of a feature for a given classification by means of the weights assigned to the feature by a linear classifier such as a support vector machine \cite{chang_feature_2008}.
      Thereby, sparsity can be emphasized by resorting to, e.g., \emph{Lasso} or sparse SVM models \cite{tibshirani_regression_1996}\cite{yao_sparse_2017}.
      Yet, provided the solution is not unique, the resulting feature relevance is to some extent arbitrary, i.e. possibly rendering the model interpretation invalid.
      Here we propose an alternative:
      instead of taking the weights of a single linear model as an approximation of feature relevances we look into using the complete model class of well-performing sparse linear classifiers.
      A model class is characterized by its similarity in the quality of the solution i.e. similar generalization ability as characterized by the size of the weight vector of the SVM~\cite{anguita_evaluating_2000} and similar training loss, which measures wrongly classified samples.
      Through the use of a class of models, we can approximate the global solution of the \emph{ARFS} problem as shown in~\cite{gopfert_interpretation_2018}.

        Assume we are interested in linear classifiers of the form 
        $\labelentry \mapsto sgn( \omega^Tx-\bias)$ where $\omega$ is the normal vector of the separating hyperplane, $\bias$ denotes the bias and \emph{sgn} refers to the sign function.
        First, we acquire a baseline solution to the problem using an $l_1$-regularized soft-margin SVM:
          \begin{align*}
            \left( \optomega,\, \optoffset,\, \optslack \right) \in &\argmin_{\omega,\, \bias,\, \slack} \onenorm{\omega} + \losssum{\slack} \\
            \text{s.t.} & \hspace{5pt} \labelentry_i(\omega^\top \sample_i  - \bias)  \geq 1 - \slack_i\,\,\, \forall i \\
            &\hspace{5pt} \slack_i                                  \geq 0 \,  \forall i   
          \end{align*}
          Through optimization, we acquire a model fully defined by the normal vector of a hyperplane $\omega$ and its offset from the origin $b$.
          Prediction of samples is based on the signed distance from the plane. 
          As usual, $\slack_i$ are slack variables to guarantee the feasibility of the constraint optimization problem in case of unavoidable classification errors.
          $\regparameter$ is a regularization parameter which depends on the datasets distribution.
          We choose the parameter guided by 3-fold stratified cross-validation and the $F_1$ score weighted by each class support to account for class imbalances.
          
          From the model with the best $\regparameter$, we obtain constraints for controlling the generalization error of equivalent models: the upper limit on the weight vector $ \mu := \optomeganorm$ and  error term $\rho := \sum^n_{i=1}{\optslack_i}$.
          These values determine the class of equivalent classifiers, which consist of all SVM solutions ($\omega^\prime$, $b^\prime$, $\slack^\prime$)  such that $\onenorm{w^\prime} \le \mu$ and $\sum\slack^\prime_i \le \rho$.
          All these alternatives are considered equivalent since they show the same performance for the given classification task as the found solution. Hence all weight vectors associated with an equivalent solution are relevant to determine the relevance of a feature to the given classification problem.

      \subsubsection{Minimum and Maximum Bounds} 
        \label{ssub:relevance_bounds}
      
        Using these constraints we now define feature relevance bounds for every feature $j$ independently i.e. we determine the interval of weight vectors which results if we take into account the weights of all possible equivalent classifiers.
        Mathematically speaking, we want to find extreme weight values for each feature given a similar error to our baseline.
        This can be done using linear programming.

        For the lower bound, the lowest possible value of feature $j$, we define the problem
          \begin{align}
            \text{minRel}&(\data,j): \min_{\omega,\, \bias,\,\slack} \abs{\omega_j} \nonumber\\
             \text{s.t.}& \hspace{5pt} \labelentry_i(\omega^\top \sample_i  - \bias)  \geq 1 - \slack_i\,,\, \slack_i \geq 0 \,\forall i \label{eq:optcons} \\
            & \sum^n_{i=1}{\slack_i} \leq \rho \nonumber \\ 
            & \onenorm{\omega} \leq (1+\delta) \cdot \mu. \nonumber
          \end{align}
          And the upper bound for $j$ is defined as
          \begin{align}
            \text{maxRel}&(\data,j): \max_{\omega,\, \bias,\, \slack} \abs{\omega_j}\nonumber \\
            \text{s.t.}& \text{ constraints } (\ref{eq:optcons}) \text{ hold.} \nonumber
          \end{align}
          The optimization problems can be rewritten as linear optimization problems and solved in polynomial time~\cite{gopfert_interpretation_2018} using appropriate solvers.
          To account for their numerical inaccuracies, which we encountered in our experiments, we also propose a relaxation factor $\delta=0.001$ in $(\ref{eq:optcons})$ to allow minor deviations from $\mu$.
          Given that all problems can be solved independently, our implementation makes use of parallel computation. 

          We end up with a real-valued matrix $\mathbb{RI} \in \IR^{2 \times d}$ which contains all pairs of relevance bounds.
          These intervals give an indication about the degree up to which a feature can or must be used in the classification and they can be visualized for interpretative purposes as seen in Fig.~\ref{fig:t21}.

        \subsubsection{Feature Classification} 
          \label{ssub:feature_classification}
          By intuition one could use the following three rules to map feature relevances to their relevance class:
            \begin{LaTeXdescription}
              \item [Strongly relevant]
              A feature is strongly relevant when its lower relevance bound is bigger than zero.
              The model class defined by its prediction accuracy, is dependent on information from it.
            \item [Weakly relevant]
              When two or more features are correlated, they can replace each other functionally in the model.
              These features are characterized by a lower bound equal to zero and an upper bound bigger than zero.
            \item [Irrelevant] 
              By definition, irrelevant features should have no measured relevance at all. Their upper (and lower) bound should therefore be zero.
            \end{LaTeXdescription}
            
            In practice, the discrimination between relevant and irrelevant features is challenging.
            The use of slack variables in the overall model and thus our relevance bounds allow variation in the contribution of features.
            For relevance bounds specifically, even if feature $j$ is independent we observe $\text{maxRel}_j>0$.
            One could introduce a naive data independent threshold to discriminate between noise and relevant features, but this would lead to bad precision or recall of features in most cases.
            Instead we try to estimate the distribution of relevances of noise features given the model constraints.
            We expect for a given model class the same amount of allowed variation in the relevances and therefore a normal distribution with an unknown mean and variance. 
            We propose to estimate the distributions parameters and the corresponding prediction interval (PI) to obtain a data dependent threshold\cite{GeisserPredictiveInference1993}.
            To estimate this noise distribution we use $n$ permutated ($p$) input features from $\data$ i.e. for each $j^p$ we compute $\text{maxRel}((\hat{x}_i,y_i)^n_{i=1},j^p)$ where $j\notin \hat{x}$.

            The prediction interval is then defined as
            \[
              PI :=\displaystyle {\overline {P}}_{n}\pm T_{n-1}(p)s_{n}{\sqrt {1+(1/n)}}.
            \]
            Here $\overline {P}_{n}$ denotes the sample mean and $s_n$ the standard deviation and $T$ represents the Student's t-distribution with $n-1$ degrees of freedom.
            The size of PI depends on parameter $p$, the expected probability that a new value is included in the interval.
            We propose default values of $p=0.999$ for a low false positive rate and $n>=50$ to achieve a stable distribution.

\begin{figure}
          \centering
          \includegraphics[width=0.5\textwidth]{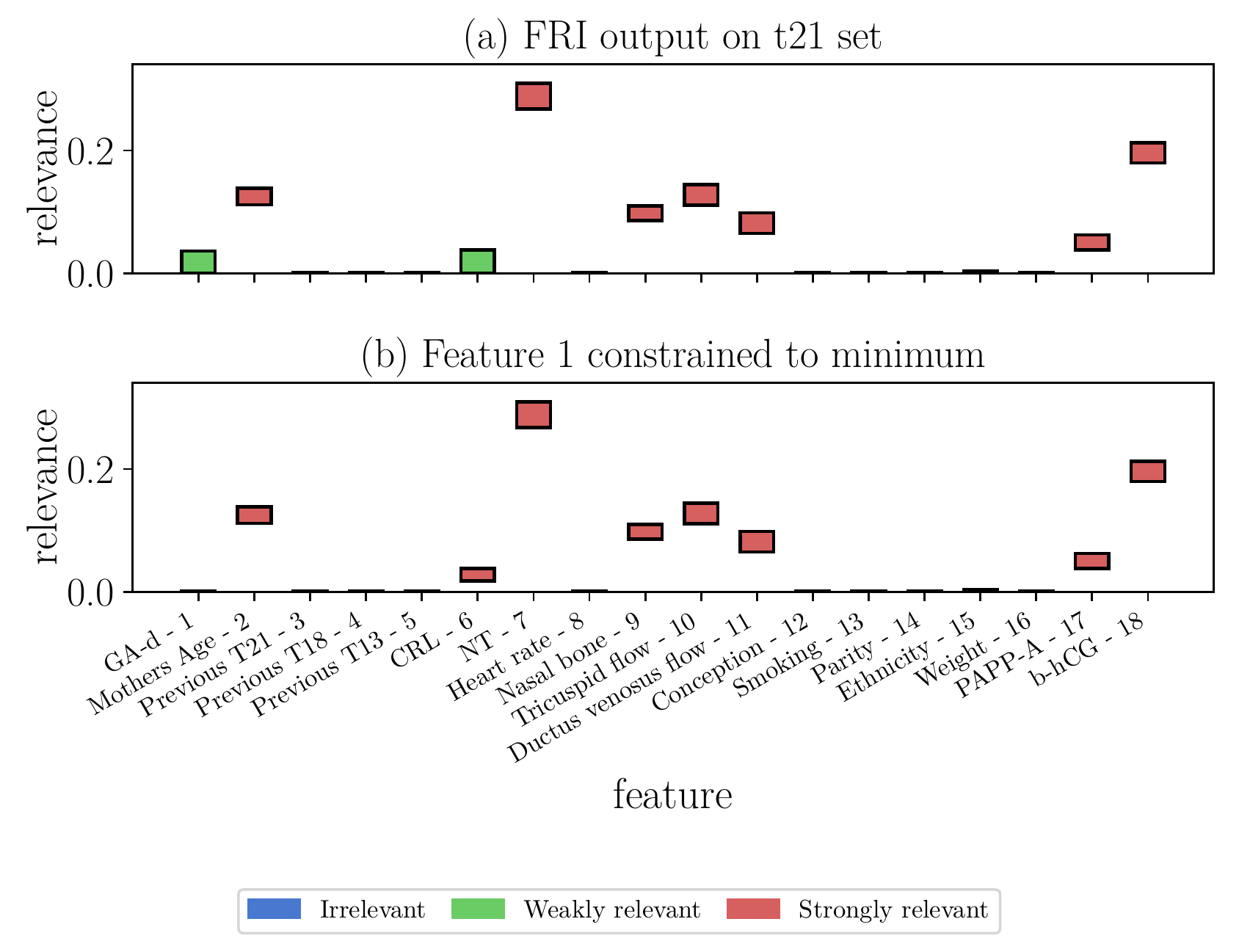}
          \caption{Program output using the \emph{t21} dataset visualizing relevance bounds for all features as colored boxes. Colors correspond to relevance classes assigned by \fri. (a) Shows program output without any constraints introduced by the user. (b) Shows output with feature 1 GA-d (``Gestation age in days'') set to its minimum value.}
          \label{fig:t21}
          \end{figure}

            To classify a feature $j$ as irrelevant we check if $\text{maxRel}_j~\in~PI$.
            To discriminate between weak and strong relevance we then additionally check the lower bound as described at the beginning of \ref{ssub:feature_classification}.
            An example of the feature classification can be seen in Fig.~\ref{fig:t21} where the relevance bars are colored according to their relevance class.
            
           Our method provides the vector $\mathbb{AR} \in \{\mathit{0},\mathit{1},\mathit{2}\}^d$, which encodes strongly $(\mathit{2})$, weakly $(\mathit{1})$ and irrelevant $(\mathit{0})$ features.

    \subsection{Feature Constraints} 
      \label{sub:feature_constraints}
          The mathematical formalization of relevance intervals as introduced above also opens up the opportunity to integrate prior knowledge about feature relevances and to interactively explore good solutions by means of integrating additional bounds for specific features.
          More precisely by solving the problem using linear programs, the addition of constraints is very easy.
          One way we leverage this is the possibility of adding relevance constraints to the optimization.

          Given data $\data$ and feature $l \in \data$ we define a set of additional constraint ranges $K$. For a constrained feature $l$ it is defined as 
          \[
            K_l:= \left[K_l^{min}, K_l^{max} \right].
          \]
          Note that $K_l\geq 0$ because relevances are by definition positive and that $|K|\neq d$, i.e. not all features have to be constrained.
          In the case of $K_l^{min} = K_l^{max}$ we consider the models usage of feature $l$ as fixed to a static value. 

          Although the values in K could be chosen arbitrarily under the given restrictions, in practice one should orientate himself using the relevance bounds in $\mathbb{RI}$.

          To compute relevance bounds including individual feature constraints, we have to extend the set of existing constraints in the optimization from Step 2 in Section~\ref{sub:relevance_optimization}.
          The minimum relevance bound with \emph{constraints} is defined as
              \begin{align}
              \label{eq:constraint}
              \text{minRel\textbf{C}}&(\mathbb{D}, j, K): \min_{\omega,\, \bias,\, \slack} \abs{\omega_j} \nonumber\\
              \text{s.t.}& \text{ constraints } (\ref{eq:optcons}) \text{ hold and} \nonumber\\
              &K_l^{min} \geq \abs{\omega_l} \geq  K_l^{max} \quad \forall l  \neq j. \nonumber
              \end{align}
          The maximum bound is defined analogously.

          To rewrite the new absolute term $\abs{\omega_l}$ as a convex problem, we utilize the baseline solution $\optomega$, which allows us to use the sign of the coefficient $\optomega_l$
          turning the non-convex absolute term into a simple convex one.

          By changing the amount of contribution allowed for one feature, we can observe varying intervals for others and infer potential dependencies between them as in Fig.~\ref{fig:t21}~(b).
          This is especially useful when designing a set of biomarkers.
          In our tool we provide the means to easily define ranges or values for all features.
          Presetting values can be based on the relevance intervals which give a first indication of importance.
          We also plan to integrate a function to automatically group features based on these constrained relevance bounds in the future.


\section{Results}

    \subsection{Feature Selection Evaluation} 
    \label{ssub:evaluation}
      To evaluate the method in context, we run benchmarks against other established methods: \emph{Boruta} as a representative of a method with statistical testing\cite{kursa_all_2011-1},
      \emph{Ensemble Feature Selection} (EFS) \cite{neumann_efs_2017}, 
      \emph{ElasticNet} using an equal contribution of $L_1$ and $L_2$ regularization  \cite{zou_regularization_2005} 
      and\emph{ stability selection} (SS)\cite{MeinshausenStabilityselection2010}\cite{ShahVariableselectionerror2013}.
      We removed the Lasso from the comparison because of very similar performance to the EN.
      For all methods, the proposed default parameters are used.
      Hyperparameters are selected according to a cross validation scheme. 
      For the ElasticNet, we choose the feature set depending on the coefficients $c_i$ of the model where $c_i>10^{-5}$ counts as selected.

      We used two types of sets: generated sets where we had knowledge of the underlying ground truth and real world data stemming from medical studies.

      \subsubsection{Simulation Data} 
        \label{sub:simulation_data}
        \label{ssub:supervised}

        All our simulation sets are sampled from a binary classification problem.
        To generate a multidimensional classification problem, we use a randomly generated prototype vector which defines a hyperplane.
        The defining features of this plane are strongly relevant.
        Now points are sampled in this feature space and the class is determined by the side of the hyperplane the points lie on.
        Weakly relevant features are constructed by replacing a feature of the original feature space with its linear combination.
        The elements of this combination are highly correlated and produce a set of redundant features.
        By removing the original feature and replacing it with those elements we achieve weak relevance by definition.
        Irrelevant features are sampled from a standard normal distribution.
        All simulation sets consist of 30 features and 500 samples.
        They differ in the density of the relevant feature space which is defined by the amount of strongly, weakly and irrelevant variables which are listed in Table~\ref{tab:simdata}.
        According to these parameters 50 sets were generated per configuration and the following evaluation refers to the aggregated scores.
        {\sone}\ and {\sthree} have a sparse relevant feature space while {\stwo} and {\sfour} are dense.
        Additionally, in {\sone} and {\stwo} weakly relevant features are present, while they are missing completely in {\sthree} and \sfour.
        {\sfive}\ had all strongly relevant features removed.

        Performance on these sets can give clues about the effectiveness of the considered feature selection strategy.
          \begin{table}[tb]
          \centering
          \caption{Characteristics of simulated datasets. Each set consists of 30 features with 500 samples.}
          \label{tab:simdata}
          \begin{tabular}{rrrr}
             \hline
            \multicolumn{1}{l}{} & \multicolumn{1}{l}{Strongly relevant} & \multicolumn{1}{l}{Weakly relevant} & \multicolumn{1}{l}{Irrelevant} \\
             \hline
            \sone                 & 4                                     & 4                                   & 22                             \\
            \stwo                & 12                                    & 8                                   & 10                             \\
            \sthree                 & 4                                     & 0                                   & 26                             \\
            \sfour                 & 18                                    & 0                                   & 12\\                          
            \sfive                 & 0                                    & 20                                   & 10\\                         
             \hline  
          \end{tabular}
          \end{table}
          Due to a known ground truth,  we can explicitly evaluate the validity of the selected features.
          We focus on the all-relevant feature selection problem and we use the following measures to evaluate the match of the detected feature set and the known ground truth of all relevant features:
          precision and recall.
          Recall is defined by TP / (TP+FN) with TP = true positives and FN = false negatives.
          It denotes how many of the relevant features were selected which is crucial when looking for the all relevant feature set.
          Precision is defined by TP / (TP+FP) with FP = false positives and describes what rate of false positives are part of the feature set.
          To get a balanced measure for a feature selection method, a combination of both is necessary.
          One can use the $F_1$ measure which is the harmonic mean of precision and recall:
          \[
            {\displaystyle F_{1}=2\cdot {\frac {\mathrm {precision} \cdot \mathrm {recall} }{\mathrm {precision} +\mathrm {recall} }}}
          \]
          The measures can be seen in Table~\ref{tbl:toyperf}.
          \begin{table*}[tb]
            \caption{Feature selection score on simulated datasets. Values are showing the performance of each method to classify between the relevance of input features.}
            \label{tbl:toyperf}
            \centering
           \begin{tabular}{l|rrrrr|rrrrr|rrrrr}
              \hline
              \textbf{score} &    F1 &       &       &       && precision &       &       &       && recall &       &       &       \\
              \textbf{data} &  Sim1 &  Sim2 &  Sim3 &  Sim4 &  Sim5 &      Sim1 &  Sim2 &  Sim3 &  Sim4 &  Sim5 &   Sim1 &  Sim2 &  Sim3 &  Sim4 &  Sim5 \\
              \hline
              \textbf{Boruta            } &  \textbf{0.98} &  0.82 &  0.91 &  0.82 &  0.98 &      0.99 &  1.00 &  0.87 &  1.00 &  1.00 &   1.00 &  0.72 &  0.98 &  0.70 &  0.95 \\
              \textbf{EFS               } &  0.96 &  0.76 &  0.71 &  0.84 &  0.94 &      0.93 &  1.00 &  0.57 &  1.00 &  1.00 &   1.00 &  0.62 &  0.98 &  0.73 &  0.90 \\
              \textbf{ElasticNet        } &  0.62 &  0.84 &  0.44 &  0.82 &  0.80 &      0.46 &  0.74 &  0.28 &  0.69 &  0.67 &   1.00 &  0.98 &  1.00 &  1.00 &  1.00 \\
              \textbf{FRI               } &  \textbf{0.98} &  \textbf{0.98} &  0.99 &  \textbf{0.99} &  \textbf{0.99} &      0.98 &  1.00 &  0.98 &  0.99 &  1.00 &   0.99 &  0.97 &  1.00 &  0.98 &  0.99 \\
              \textbf{StabilitySelection} &  0.77 &  0.75 &  \textbf{1.00} &  0.91 &  0.27 &      1.00 &  1.00 &  1.00 &  1.00 &  1.00 &   0.62 &  0.60 &  1.00 &  0.83 &  0.16 \\
              \hline
            \end{tabular}
          \end{table*}

          Before we evaluate the selection measures we confirm that all models had a proper fit.
          Listed in Table~\ref{tbl:toyaccuracy} are the training accuracies.
          One can see in the table that all classification models had accuracy values over $90\%$ which signifies a sufficient fit of the data for all of them.
            \begin{table}[tb]
            \caption{Average training set accuracy. In the case of Boruta the internal RandomForest score was reported. For \emph{EFS} accuracy is not defined. }
            \label{tbl:toyaccuracy}
            \centering
            \begin{tabular}{l|rrrrrr}
            \hline
              {} &  Boruta &  EFS &  ElasticNet &   FRI &  SS \\
              \hline
              \textbf{Sim1      } &    0.99 &  - &        1.00 &  0.92 &                1.00 \\
              \textbf{Sim2      } &    0.97 &  - &        1.00 &  0.96 &                1.00 \\
              \textbf{Sim3      } &    0.99 &  - &        1.00 &  0.96 &                1.00 \\
              \textbf{Sim4      } &    0.97 &  - &        1.00 &  0.93 &                1.00 \\
              \textbf{Sim5      } &    1.00 &  - &        1.00 &  0.91 &                1.00 \\
              \hline
              \textbf{colp.} &    1.00 &  - &        0.99 &  0.97 &                0.99 \\
              \textbf{flip      } &    1.00 &  - &        0.90 &  0.82 &                0.90 \\
              \textbf{spectf    } &    1.00 &  - &        0.99 &  0.92 &                0.98 \\
              \textbf{t21       } &    1.00 &  - &        0.98 &  0.93 &                0.98 \\
              \textbf{wbc       } &    1.00 &  - &        1.00 &  0.98 &                1.00 \\

              \hline
            \end{tabular}
            \end{table}

            To evaluate the feature selection performance we mainly observe the $F_1$ score in Table~\ref{tbl:toyperf}.
            Here our proposed \fri{} takes the lead overall with a near perfect score in all simulation sets.
            Depending on the presence of weakly relevant features, the other methods show loss of recall which leads to a reduced $F_1$ score.
            This is especially evident for \stwo{} and \sfour{} in the case of SS and EFS.
            The worst recall is achieved by SS for \sfive{} where it did not select many of the weakly relevant variables at all.
            SS still achieves slightly better scores in \sthree{} where no weakly relevant features are present.

      \subsubsection{Biomedical Data} 
        \label{sub:biomedical_data}
        The biomedical datasets are gathered from multiple studies and differ in size and type:
            \begin{itemize}
              \item \emph{t21}: This set stems from a series of prenatal examinations of pregnant women. 
              The goal of the examinations is the early diagnosis of chromosomal abnormalities, such as trisomy 21. 
              The study covers sociodemographic, ultrasonographic and serum parameters which result in 18 usable features.
              The original set contains over 50.000 samples but only a low percentage ($\approx0.8\%$) of abnormal samples.
              The data have been collected by the Fetal Medicine Centre at King’s College Hospital and University College London Hospital in London~\cite{nicolaides_multicenter_2005}.
              \item \emph{flip}:  This set is used for the prediction of fibrosis. The diagnosis of fibrosis is represented as a score which is based on sociodemographic and serum parameters.
                The set consists of 118 patients and 19 features.  
                The data were provided by the Department of Gastroenterology, Hepatology and Infectiology of the University Magdeburg~\cite{sowa_novel_2013}.
              \item \emph{spectf}: The \emph{spectf} dataset consists of 44 features describing cardiac Single Proton Emission Computed Tomography (SPECT) images.
                Each of the 267 patients images were diagnosed as either normal or abnormal.
              \item \emph{wbc}: The \emph{wbc} dataset contains 32 markers for cell image based breast cancer diagnostics from 569 patients. 
              \item \emph{colposcopy}: Set with 69 Extracted structural features from videos acquired during colposcopies \cite{FernandesTransferLearningPartial2017}. Classificiation of practitioners clinical judgment using the \emph{Schiller} modality. 
            \end{itemize}
              \emph{spectf}, \emph{wbc} and \emph{colposcopy} were acquired through the UCI Machine Learning Repository~\cite{dheeru_uci_2017}.

            The biomedical datasets are preprocessed before analysis.
            Samples with over $90\%$ missing values are removed.
            Sets are split into stratified training and testing subsets.
            If samples still contain missing feature values, we replace them with the features training set mean in both subsets.
            Similarly, the z-score transformation is based on the training set and applied to both.
            In case the original set is imbalanced, we use the Synthetic Minority Over-sampling Technique (Smote)~\cite{chawla_smote_2002} in combination with a Nearest Neighbor cleaning rule~\cite{wilson_asymptotic_1972}\cite{Laurikkala2001Improving} as described in~\cite{batista_study_2004-1}.
            In one case (\emph{t21}) with an extremely large majority class, we only perform downsampling.
            We perform the tests on 50 bootstrap replicates with sample size $0.7n$ with replacement.

            \label{ssub:unsupervised}
             To asses the quality of the feature selection on real-world datasets, we have to rely on the problem performance itself since no ground truth as regards feature relevance is available.
             We expect a FS method to pick features which contain information and a loss of features with information is signified in a decrease of performance.
             Instead of looking at each models internal accuracy score, we evaluate the selected feature sets by their discriminative power,
             whereby the latter is uniformly evaluated by a logistic regression model, which is a very popular model for predictive purposes in medical applications~\cite{bagley_logistic_2001}.
             This model is trained using only the predicted feature set and hyperparameter optimization is performed to select the model with the best cross-validated regularization parameter.
             Finally, for this selected model the Receiver operation characteristics (ROC) on the holdout validation set are recorded.
             For the comparison, we look at the area under the curve (ROC-AUC), which is listed in Table~\ref{tbl:aucreal}.
            \begin{table}[tb]
              \caption{ROC-AUC values of logistic regression model using features selected by listed models. The values are averaged over 50 bootstraps.}
              \label{tbl:aucreal}
              \centering
                \begin{tabular}{l|rrrrrr}
                \hline
                 & Boruta &  EFS &  EN &   FRI &  SS  \\
                \hline
                \textbf{colposcopy} &  0.568 &  0.586 &      0.640 &  \textbf{0.661} &              0.625 \\
                \textbf{flip      } &  0.804 &  0.652 &      \textbf{0.815} &  0.743 &              0.705 \\
                \textbf{spectf    } &  0.871 &  0.874 &      0.867 &  0.880 &              \textbf{0.888} \\
                \textbf{t21       } &  0.971 &  0.977 &      0.971 &  0.975 &              \textbf{0.978} \\
                \textbf{wbc       } &  0.997 &  0.998 &      0.998 &  0.998 &              0.999 \\
                \hline
                \end{tabular}
              \end{table}
              Here the AUC on the five datasets produces no clear overall superior method which is in line with the expectation that the minimal optimal set is the objective of most methods and sufficient for prediction.
              On the \emph{spectf}, \emph{t21} and \emph{wbc} datasets most methods produce very similar performing feature sets.
              In the case of \emph{colposcopy} the feature set selected by FRI achieves the best performance.
              SS produces sightly better sets in two cases.
              The ElasticNet performs solid in all cases based on its very conservative selection method where informative features are not removed often.

              This leads the part of the evaluation more concerned with the goal stated in the introduction.
              In the search for an interpretative and complete feature set we need to take the selected set size into account.
              Table~\ref{tbl:mean_set_size} lists the average feature set sizes over all experiments.
              Because \fri{} provides additional information by not only conserving all weakly relevant features but also by denoting the feature class itself we can explicitly list those as well.
              As mentioned in the last paragraph, we can easily see that EN is very conservative in its selection.
              It produces by far the biggest feature sets with many false positives in the case of the \emph{Sim} sets but also most likely in the real datasets.
              Similarly for Boruta, which achieves better precision in the generated data but still shows seemingly inflated set sizes.
              SS on the other hand exhibits very good precision overall.
              Interestingly the size of the sets chosen by \emph{stability selection} is very similar to FRI$_s$, the set of strongly relevant features chosen by \fri{}.
              This indicates that \fri{} can find strongly relevant features with high precision, but also highlights the additional information provided by the weakly relevant features contained in FRI$_w$.

              But why do we need the information in FRI$_w$?
              
              \begin{table}[tb]
                \caption{Average selected feature set size. Additionally for \emph{FRI} the size of the strongly($_s$) and weakly ($_w$) relevant feature set is available.}
                \label{tbl:mean_set_size}
                \centering
                    \begin{tabular}{lrrrrr||rr}
                    \hline
                     &  Boruta &   EFS &  EN &  SS &   FRI &  FRI$_s$ &  FRI$_w$ \\
                    \hline
                      \textbf{Sim1      } &     8.1 &   8.7 &        17.8 &                 5.0 &   8.1 &    5.1 &    3.0 \\
                      \textbf{Sim2      } &    14.3 &  12.3 &        26.6 &                12.1 &  19.4 &   12.4 &    7.0 \\
                      \textbf{Sim3      } &     4.6 &   7.2 &        14.8 &                 4.0 &   4.1 &    4.0 &    0.1 \\
                      \textbf{Sim4      } &    12.6 &  13.2 &        26.2 &                15.0 &  17.9 &   17.9 &    0.0 \\
                      \textbf{Sim5      } &    19.1 &  17.9 &        29.7 &                 3.2 &  19.9 &    0.0 &   19.9 \\
                      \hline
                      \textbf{colp.} &    35.1 &  25.4 &        46.5 &                41.5 &  20.3 &    5.9 &   14.4 \\
                      \textbf{flip      } &    18.8 &   8.1 &        16.9 &                 9.1 &   8.9 &    8.8 &    0.1 \\
                      \textbf{spectf    } &    44.0 &  20.3 &        43.1 &                 5.9 &  19.9 &    5.9 &   14.0 \\
                      \textbf{t21       } &    15.5 &   7.9 &        14.2 &                 9.6 &   9.6 &    6.6 &    3.0 \\
                      \textbf{wbc       } &    29.9 &  12.5 &        26.9 &                 4.7 &  15.6 &    4.0 &   11.6 \\
                    \hline
                    \end{tabular}
               \end{table}
      \subsection{Evaluation of Interactive Use} 
      \label{sub:example_of_interactive_use}
          By having additional information available in the set of weakly relevant features FRI$_w$
          we can gain insights into the structure of the data.
          We can improve the design of models and diagnostic tests in biomedical applications.
          Our framework given in \ref{sub:feature_constraints} allows introducing constraints into the model.
          This makes it possible to limit the contribution of certain features to specific intervals or a fixed value.
          These limits can come from prior knowledge of the practitioner and represent design goals or existing hypotheses.
          Depending on the chosen values the model and the resulting relevance bounds change and can be visualized again which lends itself to an iterative and interactive process.
          In the following, we are going to evaluate that use case on simulated data and on the \emph{t21} data set.

\begin{figure}[tb]
          \begin{center}
            \includegraphics[width=0.35\textwidth]{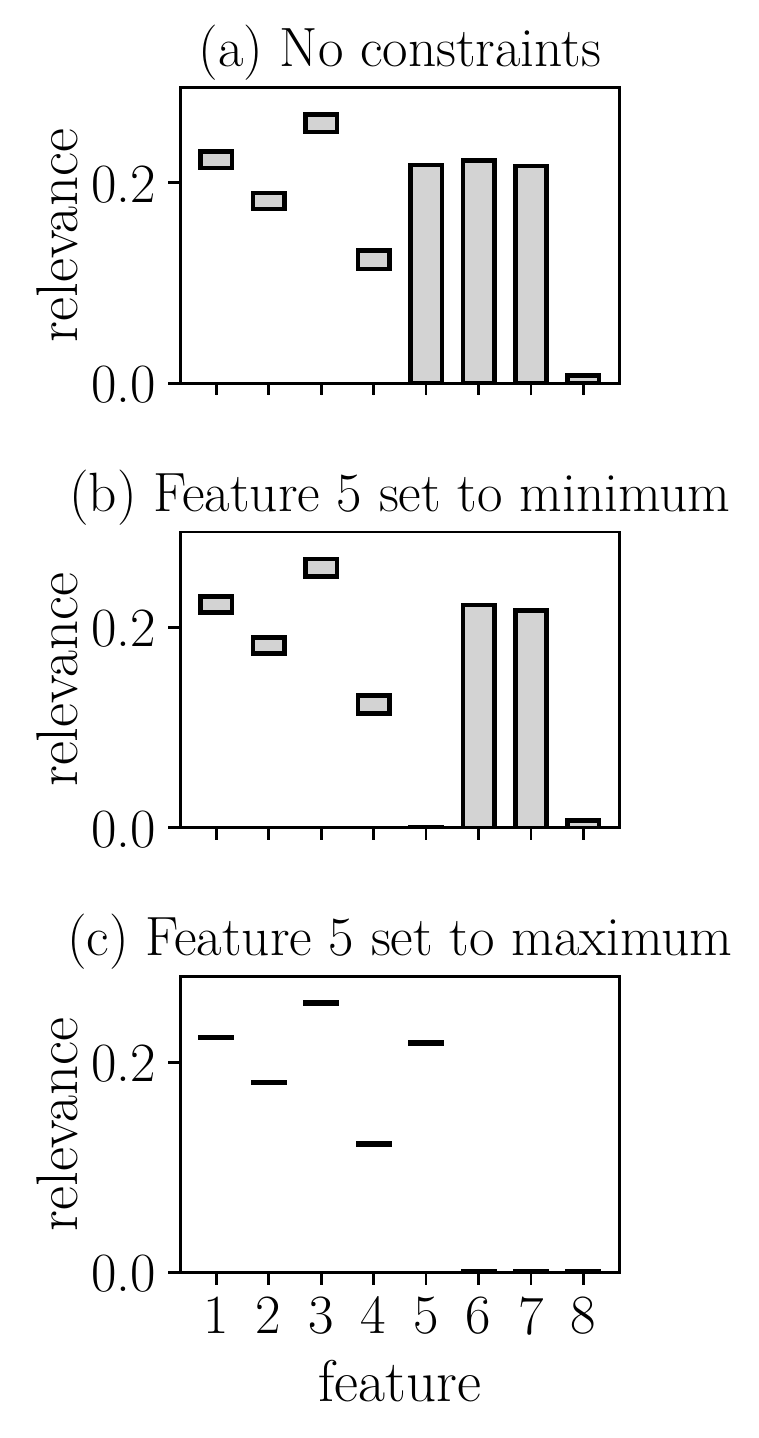}
          \end{center}
          \caption{Three subplots showing feature relevance bounds in different constraint situations according to section \ref{sub:feature_constraints}. Classification data were simulated and consisted of 4 strongly relevant features (1-4), 3 weakly relevant (5-7) and one noise feature (8). Subplot (a) had no feature constraints. Subplot (b) shows the output when feature 5 is constrained to its minimum relevance value and (c) to its maximum value.}
          \label{fig:inter_sim}
        \end{figure}

          The simulated set was generated according to \ref{sub:simulation_data}.
          It consists of 8 features, 4 of which are strongly relevant, 3 of which are weakly relevant and one noise feature.
          Fig.~\ref{fig:inter_sim} (a) shows the output of \fri{} without any constraints.
          The four strongly relevant features (1-4) are visible as four small rectangles with lower relevance bounds (the bottom part of the rectangle) bigger than zero.
          The model parameters allow some variation in their contribution to the model.
          The three weakly relevant features (5-7) are visible as three taller rectangles with equal height because they are perfectly correlated in the normalized space.
          They can replace each other in the model.
          This is apparent when we preset one of them (e.g feature 5) to the minimum and maximum relevance bound 
          i.e. we calculate $\text{minRel\textbf{C}}(\mathbb{D},j,K)$ and $\text{maxRel\textbf{C}}(\mathbb{D},j,K)$
          for all $j\neq 5$ and $K=\{K_5\}$. $K_5$ is then either $\mathbb{RI}^{min}_5$ or $\mathbb{RI}^{max}_5$, i.e fixed to a static value.

          In  Fig.~\ref{fig:inter_sim} (b) feature 5 was set to the minimum bound and the relevance bounds of other features are identical.
          That is because the other two features are still a correlated pair which allows the same degree of variability in contribution.
          When feature 5 is set to its maximum relevance bound in (c) we see that feature 6 and 7  have no contribution anymore.
          Additionally, all other relevance bounds are reduced to single values because the model in this state does not allow any more variability.

          After this experiment, it is now interesting to apply this procedure on real data with functional associations between features.
          In Fig.~\ref{fig:t21} (a) the normal \fri{} output of the \emph{t21} set is presented.
          As a reminder, this set consists of samples acquired in prenatal examinations of mothers and their unborn children.
          Features in the study included socioeconomic factors as well as ultrasound imaging metrics.
          Notably in the output of \fri{} are two weakly relevant features 1 and 6.
          Feature 1 represents the gestational age of the fetus in days (`GA-d') and feature 6 the crown rump length (`CRL') of the fetus, which is the length as indicated on an ultrasound machine.
          By intuition, we expect an association between the two measures.
          If we set one of the two features to its minimum relevance bound (Fig.~\ref{fig:t21} (b)) we see that feature 6 becomes strongly relevant in the model.
          This highlights the association between the two which is very useful in cases where it is not clear a priori.
          Furthermore, we can use this as a design tool to easily select `better' features.
          If we find functional alternatives, we can exclude more expensive features in future experiments or tests.         


      \subsection{Runtime} 
      \label{sub:runtime}
        The computational runtime of a method is not only an indicator for its feasibility on bigger datasets 
        but also especially important in the interactive use case where an analyst is actively involved in model refinement.
        Because relevance bounds can be solved independently in parallel, we provide the means to speed up computation by utilizing all available CPU cores on the machine.
        Additionally, we also tested running our program in conjunction with the distributed 
        computation framework \emph{Dask} which allows scaling up to any amount of separate computing nodes in a high performance cluster such as \emph{Grid Engine} or even in cloud backends.
        
\begin{figure}[tb]
            \centering
            \includegraphics[width=0.5\textwidth]{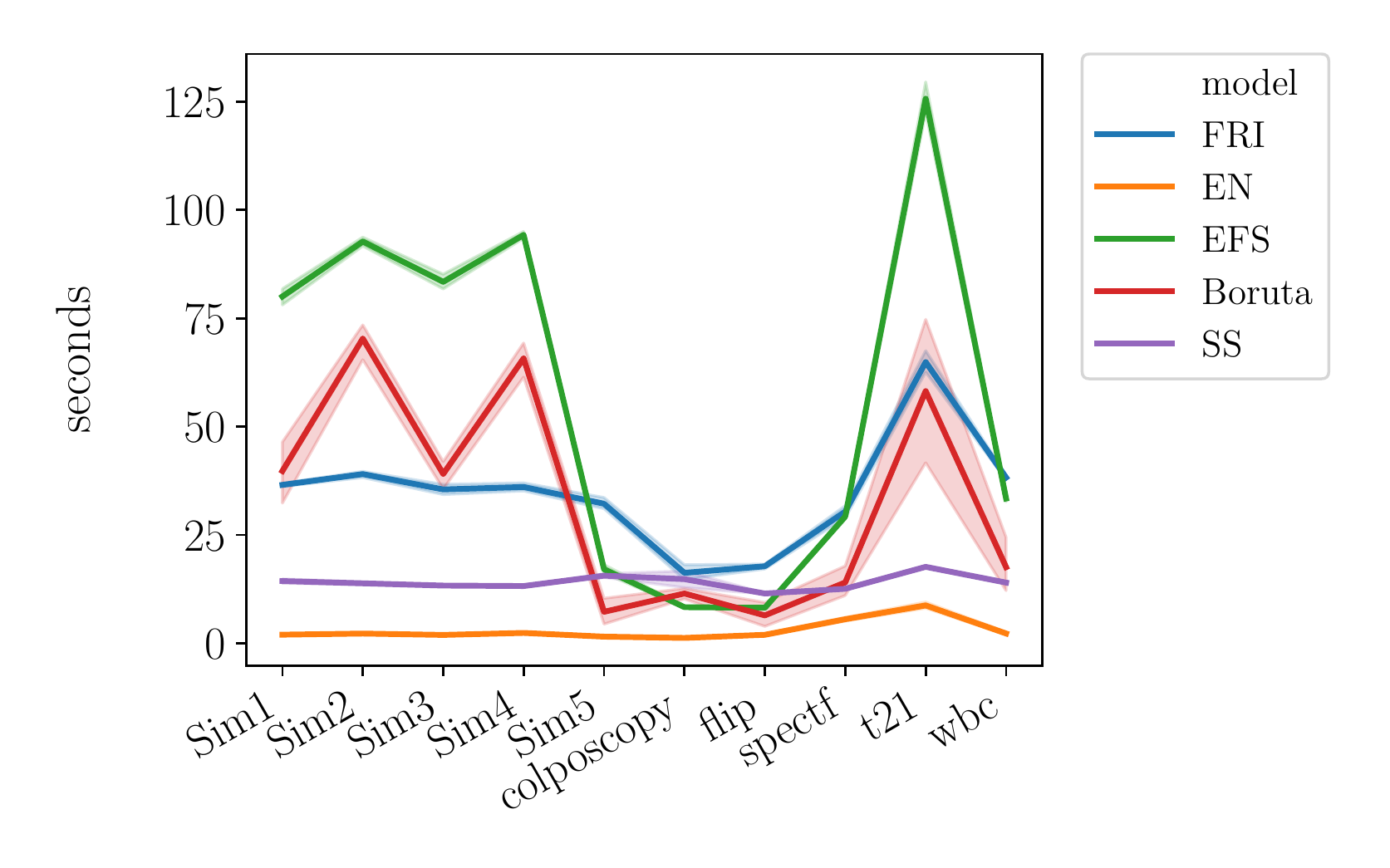}
            \caption{Average runtime over all bootstraps with confidence intervals.}
            \label{fig:runtime}
        \end{figure}

        In Fig.~\ref{fig:runtime} we display aggregated mean runtime of the methods used in our evaluations on a single CPU thread.
        Because only \fri\ and one other method provided a parallel implementation (SS) the computations were limited to one thread, so an advantage of the parallel processing is not taken into account.
        EN performed best followed by SS.
        Both show steady runtimes over all types of data.
        \emph{Boruta’s} runtime is very dependent on the density of the feature space and shows some variance in the case of \emph{t21}.
        The runtime of \fri\ is similar to \emph{Boruta} in most cases but takes a hit in smaller datasets because of the constant factor of sampling permutated features for feature classification. 
        \emph{EFS} shows the slowest performance in most cases, which clearly stems from its use of multiple complex underlying models at the same time.

\section{Conclusion}
  We have presented the software library \fri\ to produce all relevant feature sets for general feature selection as well as 
  perform interactive data exploration.
  We described how we implemented the algorithm from~\cite{gopfert_interpretation_2018} and extended the method to allow a practitioner to include new constraints and experiment.
  We also proposed a threshold estimation method to reduce false positives which are common in all-relevant selection tasks.

  In comparison with other methods, we showed that \fri\ can detect all relevant features in synthetic datasets while minimizing noise through its threshold estimation.
  On real datasets we showcased good selection performance and additional information provided by the weakly relevant feature set.
  Our underlying method ensures to conserve all relevant variables while still maintaining interpretability.
  This is facilitated by the three relevance classes our method produces as well as the relevance bar representation
   which should enable better understanding for biological and medical experts in the future.
  In addition to facilitating understanding we also provide a way to incorporate prior knowledge to manipulate the model itself which should help in the design of new experiments and biomarkers for prediction models.
  
\section*{Acknowledgments}
The authors would like to thank Professor Kypros Nicolaides
and Dr. Argyro Syngelaki from Fetal Medicine Foundation
for making available the \emph{t21} data used in this study.
We are also grateful for the \emph{flip} datasets provided by Professor Ali Canbay, Department of Gastroenterology, Hepatology, and Infectiology of the University Hospital Magdeburg.

\bibliographystyle{vancouver}
\bibliography{cibcb}
\end{document}